\documentclass[10pt,twocolumn,letterpaper]{article}

\usepackage{cvpr}
\usepackage{times}
\usepackage{epsfig}
\usepackage{graphicx}
\usepackage{amsmath}
\usepackage{amssymb}
\usepackage{subcaption}
\usepackage{makecell}
\usepackage{multirow}
\usepackage[export]{adjustbox}
\usepackage{mwe}
\usepackage{enumitem}
\usepackage{mmstyle}
\usepackage{tablefootnote}
\usepackage[numbers]{natbib}

\usepackage[pagebackref=true,breaklinks=true,letterpaper=true,colorlinks,bookmarks=false]{hyperref}
\graphicspath{{figures/},{imagenet/}}

\cvprfinalcopy 

\def\cvprPaperID{801} 

\ifcvprfinal\fi

\def\imw#1#2{\includegraphics[width=#2\linewidth]{#1}}

\def\imwh#1#2#3{\includegraphics[width=#2\linewidth,height=#3\textheight]{#1}}

\newcommand{\tb}[3]{\setlength{\tabcolsep}{#2mm}\begin{tabular}{#1}#3\end{tabular}}

\begin{document}

\title{Unsupervised Feature Learning via Non-Parametric Instance Discrimination}

\author{
	Zhirong Wu$^{\star\dagger}$~~~~~~~~~~Yuanjun Xiong$^{\dagger\ddagger}$~~~~~~~~~~Stella X. Yu$^{\star}$~~~~~~~~~~Dahua Lin$^{\dagger}$ \\
	$^\star$UC Berkeley / ICSI~~~~~~~~~$^\dagger$Chinese University of Hong Kong~~~~~~~~~$^\ddagger$Amazon Rekognition}

\maketitle

\begin{abstract}

%
Neural net classifiers trained on data with annotated class labels can also capture apparent visual similarity among categories without being directed to do so.  We study 
whether this observation can be extended beyond
the conventional domain of supervised learning: 
Can we learn a good feature representation that captures apparent similarity among {\it instances}, instead of classes, by merely asking the feature to be discriminative of individual instances?

We formulate this intuition as a non-parametric classification
problem at the instance-level, and use noise-contrastive estimation
to tackle the computational challenges imposed by the large number of instance classes.
%

Our experimental results demonstrate that, 
under unsupervised learning settings,
our method surpasses the state-of-the-art on ImageNet classification by a large margin.
Our method is also remarkable for consistently improving test performance with more training data and better network architectures.
By fine-tuning the learned feature, we further obtain competitive results for semi-supervised learning and object detection tasks.
Our non-parametric model is highly compact: With $128$ features per image, our method
requires only $600$MB storage for a million images, enabling fast nearest neighbour retrieval at the run time.

\end{abstract}

\section{Introduction}
\label{introduction}

\def\row#1{
\imwh{imagenet/leopard/#1.jpg}{0.16}{0.05}&
\imwh{imagenet/jaguar/#1.jpg}{0.16}{0.05}&
\imwh{imagenet/cheetah/#1.jpg}{0.16}{0.05}&
\imwh{imagenet/boat/#1.jpg}{0.16}{0.05}&
\imwh{imagenet/cart/#1.jpg}{0.16}{0.05}&
\imwh{imagenet/bookcase/#1.jpg}{0.16}{0.05}\\
}
\begin{figure}[t]
        \centering
\imw{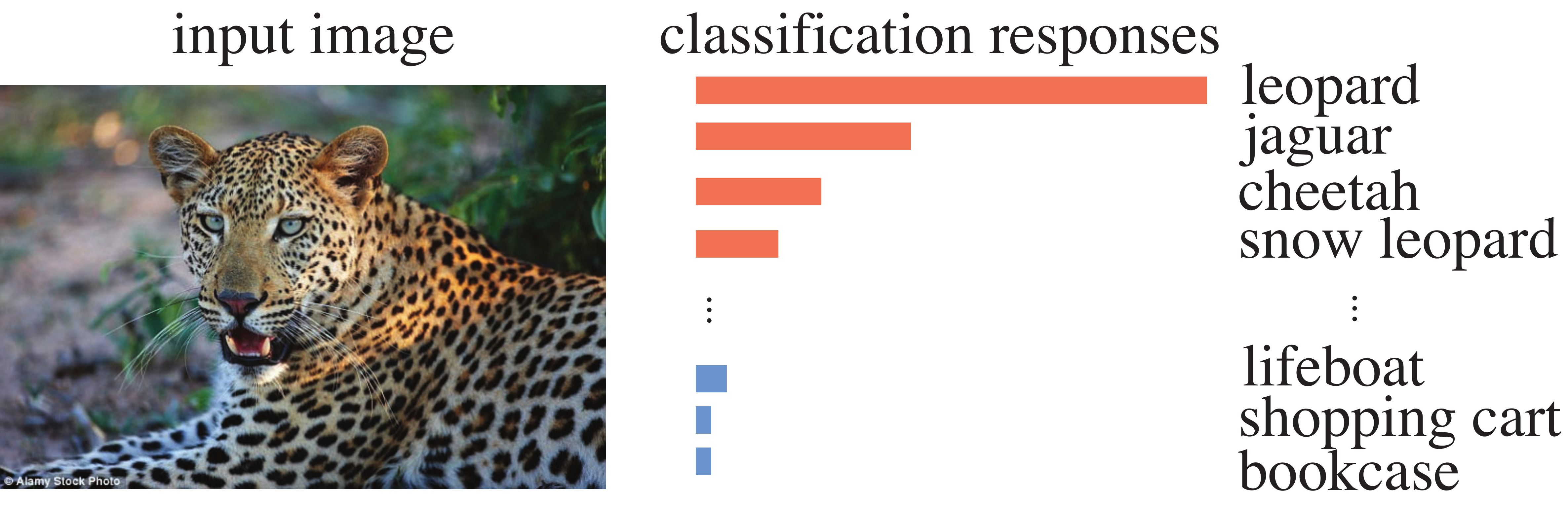}{1}\\
{\small
\tb{@{}ccc@{\hspace{2mm}}ccc@{}}{0.2}{
\row{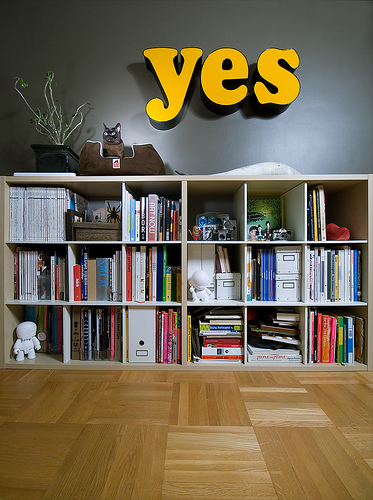}
\row{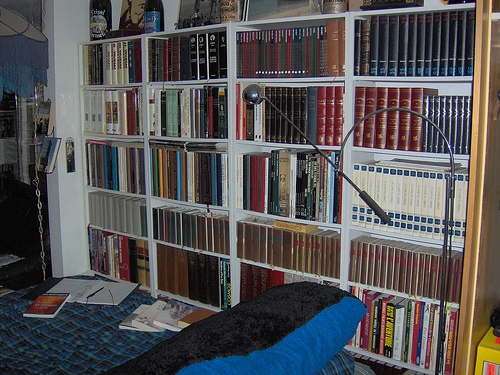}
\row{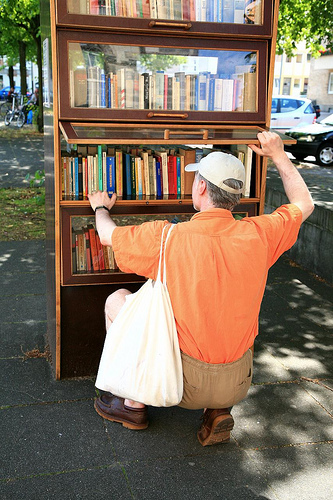}
leopard&
jaguar&
cheetah&
lifeboat&
shopcart&
bookcase\\
}}\\
\caption{\small Supervised learning results that motivate our unsupervised approach.  For an image from
class {\it leopard}, the classes that get highest responses from a trained neural net classifier are all visually correlated, e.g., {\it jaguar} and {\it cheetah}.
It is not the semantic labeling, but the apparent similarity in the data themselves that brings some classes closer than others.
Our unsupervised approach takes the class-wise supervision to the extreme and learns a feature representation that discriminates among individual instances.
}
\label{fig:teaser}
\end{figure}



The rise of deep neural networks, especially convolutional neural networks (CNN),
has led to several breakthroughs in computer vision benchmarks.
Most successful models are trained via supervised learning, which requires large datasets that are completely annotated for a specific task.
However, obtaining annotated data is often very costly or even infeasible in certain cases. In recent years, unsupervised learning has received increasing attention from the community~\cite{dosovitskiy2014discriminative, doersch2015unsupervised}.


Our novel approach to unsupervised learning stems from a few observations on the results of supervised learning for object recognition.
On ImageNet, the top-5 classification error is
significantly lower than the top-1 error~\cite{krizhevsky2012imagenet}, and
the second highest responding class in the softmax output to an image is more likely to be visually correlated.  Fig.~\ref{fig:teaser} shows that
  an image from class {\it leopard} 
is rated much higher by class {\it jaguar} rather than by class {\it bookcase}~\cite{hinton2015distilling}.
Such observations reveal that a typical discriminative learning method can automatically discover apparent  similarity among semantic categories, without being explicitly guided to do so.
In other words, apparent similarity is learned not from semantic annotations, but from the visual data themselves.



We take the class-wise supervision to the extreme of instance-wise supervision, and ask:
Can we learn a meaningful metric that reflects apparent similarity among {\it instances}  via pure discriminative learning?
%
An image is distinctive in its own right,
and each could differ  significantly from other images in the same semantic category~\cite{malisiewicz2011ensemble}.
If we learn to discriminate between individual instances, without any notion of semantic categories, we may end up with a representation that captures apparent similarity \emph{among instances}, just like how class-wise supervised learning still retains apparent similarity \emph{among classes}.
This formulation of unsupervised learning as an instance-level discrimination is also technically appealing, as it could
benefit from latest advances in discriminative supervised learning, \eg~ on new network architectures.


However, we also face a major challenge, now that the number of \emph{``classes''} is the size of the entire training set.  For ImageNet, it would be 1.2-million instead of 1,000 classes.
Simply extending softmax to many more classes becomes infeasible.
We tackle this challenge by approximating the full softmax distribution
with noise-contrastive estimation (NCE)~\cite{gutmann2010noise},
and by resorting to a proximal regularization method~\cite{parikh2014proximal} to stabilize
the learning process.

To evaluate the effectiveness of unsupervised learning, 
past works such as ~\cite{doersch2015unsupervised,pathak2016context} have relied on a linear classifier, \eg~Support Vector Machine (SVM), to connect the learned feature to categories for classification at the test time.
However, it is unclear
why features learned via a training task
could be \emph{linearly} separable for an unknown testing task.


We advocate a non-parametric approach for both training and testing.  We formulate instance-level discrimination as a metric learning problem, where distances (similarity) between instances are calculated directly from the features in a non-parametric way.  That is, the features for each instance are stored in a discrete memory bank, rather than weights in a network.
At the test time, we perform classification using
\emph{k}-nearest neighbors (kNN) based on the learned metric.
Our training and testing are thus consistent, since both learning and evaluation of our model are concerned with the same metric space between images.
We report and compare experimental results with both SVM and kNN accuracies.

Our experimental results demonstrate that, under unsupervised learning settings, our method surpasses the state-of-the-art on image classification  by  a  large  margin,  with top-1 accuracy $46.5\%$ on
ImageNet 1K~\cite{deng2009imagenet}  and  $41.6\%$ for Places 205~\cite{zhou2014learning}.
Our method is also remarkable for consistently improving test performance with more training data and better network architectures.
By fine-tuning the learned feature, we further obtain competitive results for semi-supervised learning and object detection tasks.
Finally, our non-parametric model is highly compact: With $128$ features per image, our method
requires only $600$MB storage for a million images, enabling fast nearest neighbour retrieval at the run time.



\section{Related Works}
\label{relatedworks}

There has been growing interest in unsupervised learning
without human-provided labels.
Previous works mainly fall into two categories:
1) generative models and 2) self-supervised approaches.

\vspace{3pt}

\noindent \textbf{Generative Models. }
The primary objective of generative models is to reconstruct the distribution of data as faithfully as possible.
Classical generative models include
Restricted Bolztmann Machines (RBMs)~\cite{hinton2006fast, tang2012robust, lee2009convolutional}, and
Auto-encoders~\cite{vincent2008extracting, le2013building}.
The latent features produced
by generative models could also help object recognition.
Recent approaches such as generative adversarial networks~\cite{goodfellow2014generative, donahue2016adversarial}
and variational auto-encoder~\cite{kingma2013auto} improve
both generative qualities and feature learning.

\vspace{3pt}

\noindent \textbf{Self-supervised Learning.}
Self-supervised learning exploits internal structures of data
and formulates \emph{predictive} tasks to train a model.
Specifically, the model needs to predict either an omitted aspect or component of an instance given the rest.
To learn a representation of images, the tasks could be:
predicting the context~\cite{doersch2015unsupervised},
counting the objects~\cite{noroozi2017representation},
filling in missing parts of an image~\cite{pathak2016context},
recovering colors from grayscale images~\cite{zhang2016colorful}, or
even solving a jigsaw puzzle~\cite{noroozi2016unsupervised}.
For videos, self-supervision strategies include: leveraging 
temporal continuity via tracking~\cite{wang2015unsupervised, wang2017transitive},
predicting future ~\cite{walker2016uncertain},
or preserving the equivariance of
egomotion~\cite{jayaraman2017learning, zhou2017unsupervised, pathak2016learning}.
Recent work~\cite{doersch2017multi} attempts to combine several
self-supervised tasks to obtain better visual representations.
Whereas self-supervised learning may capture relations among parts or aspects
of an instance, it is unclear why a particular self supervision task should help semantic recognition and which task would be optimal.

\vspace{3pt}

\noindent \textbf{Metric Learning.}
Every feature representation $F$  induces a metric between instances $x$ and $y$: $d_F(x, y) = \|F(x) - F(y)\|$.
Feature learning can thus also be viewed as a certain form of
metric learning.
There have been extensive studies on
metric learning~\cite{koestinger2012large, roweis2004neighbourhood}.
Successful application of metric learning can often result in
competitive performance, \eg~on face recognition~\cite{schroff2015facenet} and
 person re-identification~\cite{xiao2017joint}.
In these tasks, the classes at the test time are disjoint from those at the training time.  Once a network is trained, one can only infer from its feature representation, not from the subsequent linear classifier.  Metric learning has been shown to be effective for few-shot learning~\cite{sohn2016improved, vinyals2016matching,
snell2017prototypical}.
An important technical point 
on metric learning for face recognition is normalization ~\cite{schroff2015facenet,liu2017sphereface,wang2017normface},
which we also utilize in this work.
Note that all the methods mentioned here require supervision in certain ways. Our work is drastically different: It
learns the feature and thus the induced metric in an \emph{unsupervised} fashion, without any human annotations.

\vspace{3pt}

\begin{figure*}[t]
	\centering
	\includegraphics[width=0.98\textwidth]{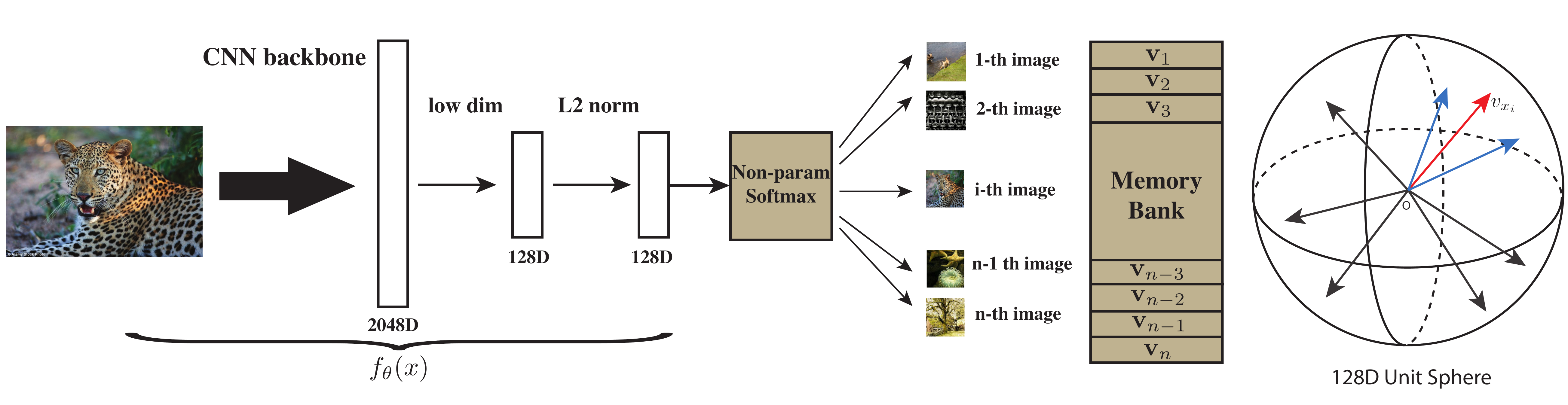}
	\caption{\small
	The pipeline of our unsupervised feature learning approach.
	We use a backbone CNN
	to encode each image as a feature vector, which
	is projected to a $128$-dimensional space and L2 normalized.
	The optimal feature embedding is learned via instance-level discrimination,
	which tries to maximally scatter the features of training samples	over the 128-dimensional unit sphere.}
	\label{fig:pipeline}
\end{figure*}

\noindent \textbf{Exemplar CNN.}
\emph{Exemplar CNN}~\cite{dosovitskiy2014discriminative} appears similar to our work.
The fundamental difference is that it
adopts a parametric paradigm during both training and testing,
while our method is non-parametric in nature.
We study this essential difference experimentally in Sec \ref{p:np}.
\emph{Exemplar CNN} is computationally demanding for large-scale datasets such as
ImageNet.

%



\section{Approach}
\label{LFD}

Our goal is to learn  an embedding function $\vv = f_{\vtheta}(x)$ without supervision.
$f_{\vtheta}$ is a deep neural network
with parameters $\vtheta$, mapping image $x$ to
feature $\vv$.
This embedding would induces a metric over the image space, as
$d_{\vtheta}(x, y) = \|f_{\vtheta}(x) - f_{\vtheta}(y)\|$ for instances $x$ and $y$.
A good embedding should map visually similar images closer to each other.

Our novel unsupervised feature learning approach is
\emph{instance-level discrimination}.
We treat
\emph{each image instance as a distinct class of its own}
and train a classifier to distinguish between  individual instance classes (Fig.\ref{fig:pipeline}).

\subsection{Non-Parametric Softmax Classifier}

\noindent \textbf{Parametric Classifier.}
We formulate the instance-level classification objective using the softmax criterion. 
Suppose we have $n$ 
images $x_1, \ldots, x_n$ in $n$ classes and their
features $\vv_1, \ldots, \vv_n$ with $\vv_i = f_{\vtheta}(x_i)$.
Under the conventional parametric softmax formulation,
for image $x$ with feature
 $\vv = f_{\vtheta}(x)$,
the probability of it being recognized as $i$-th example is
\begin{equation}\label{eqn:p_softmax}
	P(i | \vv) =
	\frac{\exp \left( \vw_i^T \vv \right)}
	{\sum_{j=1}^n \exp \left( \vw_j^T \vv \right)}.
\end{equation}
where $\vw_j$ is a weight vector for class $j$, and $\vw_j^T \vv$ 
measures how well $\vv$ matches the $j$-th class \ie,~instance.

\noindent \textbf{Non-Parametric Classifier.}
The problem with the parametric softmax formulation in Eq.~(\ref{eqn:p_softmax})
is that the weight vector $\vw$ serves as a class prototype, preventing explicit comparisons between instances.

We propose a \emph{non-parametric} variant of Eq.~(\ref{eqn:p_softmax}) that replaces
$\vw_j^T \vv$ with $\vv_j^T \vv$, and we enforce $\|\vv\| = 1$
via a L2-normalization layer.
%
Then the probability $P(i | \vv)$ becomes:
\begin{equation}\label{eqn:np_softmax}
	P(i | \vv) =
	\frac{\exp \left( \vv_i^T \vv / \tau \right)}
	{\sum_{j=1}^n \exp \left( \vv_j^T \vv / \tau \right)},
\end{equation}
where $\tau$ is a temperature parameter that controls the concentration level of the distribution~\cite{hinton2015distilling}.
$\tau$ is important for supervised feature learning ~\cite{wang2017normface}, and also necessary for tuning the concentration of $\vv$ on our unit sphere.

The learning objective is then to maximize the joint probability
$\prod_{i=1}^n P_{\vtheta}(i | f_{\vtheta}(x_i))$, 
or equivalently to
minimize the negative log-likelihood over the training set, as
\begin{equation} \label{eqn:obj0}
J(\vtheta)
=  - \sum_{i=1}^n \log P(i | f_{\vtheta}(x_i)).
\end{equation}

\noindent \textbf{Learning with A Memory Bank.}
To compute the probability $P(i|\vv)$ in Eq.~\eqref{eqn:np_softmax}, 
 $\{\vv_j\}$ for all the images are needed.
Instead of exhaustively computing these representations every time,
we maintain a feature memory bank $V$ for storing them ~\cite{xiao2017joint}.
In the following, we introduce separate notations for the memory bank and features forwarded from the network.
Let $V = \{\vv_j\}$ be the memory bank and $\vf_i = f_{\vtheta}(x_i)$ be the feature of $x_i$.
During each learning iteration, the representation $\vf_i$ as well as the network parameters $\theta$ are optimized via
stochastic gradient descend.
Then $\vf_i$ is updated to $V$ at the corresponding instance entry $\vf_i \to \vv_i$.
We initialize all the representations in
the memory bank $V$ as unit random vectors.

\noindent \textbf{Discussions.}
The conceptual change from class weight vector $\vw_j$ to feature representation $\vv_j$ directly is significant.
The weight vectors $\{\vw_j\}$ in the original
softmax formulation are only valid for training classes.  Consequently, they are not generalized to new classes, or in our setting, new instances.
When we get rid of these weight vectors, 
our learning objective focuses entirely on the feature representation and its induced \emph{metric}, which can be applied everywhere in the space and to any new instances at the test time.

Computationally, our non-parametric formulation
eliminates the need for computing and storing
the gradients for $\{\vw_j\}$, making it more scalable for big data applications.

\subsection{Noise-Contrastive Estimation}
\label{sub:nce}
Computing the non-parametric softmax in Eq.\eqref{eqn:np_softmax}
is cost prohibitive when the number of classes $n$ is very large, e.g. at the scale of millions.
Similar problems have been well addressed in the literature for learning
word embeddings~\cite{mnih2013learning,mikolov2013distributed}, where the number
of words can also scale to millions. Popular techniques to reduce computation
include hierarchical softmax~\cite{morin2005hierarchical}, noise-contrastive
estimation (NCE) ~\cite{gutmann2010noise}, and negative
sampling~\cite{mikolov2013distributed}. We use
NCE ~\cite{gutmann2010noise} to approximate the full softmax.

We adapt NCE to our problem, in order to tackle the difficulty of
computing the similarity to all the instances in the training set.
%
The basic idea is to cast the multi-class classification problem into a set of
binary classification problems, where the binary classification task is to discriminate between 
\emph{data samples} and \emph{noise samples}.
Specifically, the probability that feature representation $\vv$ in the memory bank 
corresponds to the $i$-th example under our model is,
\begin{align} \label{eqn:Z}
P(i | \vv) & = \frac{\exp(\vv^T \vf_i / \tau)}{Z_i}\\
Z_i = &\sum_{j=1}^n \exp \left( \vv_j^T \vf_i / \tau \right)
\end{align}
where $Z_i$ is the normalizing constant.
We formalize the \emph{noise distribution}
as a uniform distribution: $P_n = 1/n$.
Following prior work, we assume that noise samples are $m$ times more
frequent than data samples.
Then the posterior probability of sample $i$ with feature $\vv$ being
from the data distribution (denoted by $D = 1$) is:
\begin{equation}
h(i, \vv) := P(D=1 | i, \vv) = \frac{P(i|\vv)}{P(i|\vv) + m P_n(i)}.
\end{equation}
Our approximated training objective is to minimize
the negative log-posterior distribution of data and noise samples,
\begin{align}
J_{NCE}(\vtheta)
=  - &E_{P_d} \left[\log h(i, \vv)\right] \notag\\
   - m \cdot & E_{P_n} \left[\log (1 - h(i, \vv')) \right].
\end{align}
Here, $P_d$ denotes the actual data distribution.
For $P_d$, $\vv$ is the feature corresponding to $x_i$;
whereas for $P_n$, $\vv'$ is the feature from another image, randomly sampled according to
noise distribution $P_n$.
In our model, both $\vv$ and $\vv'$ are sampled from the non-parametric memory bank $V$. 

Computing normalizing constant $Z_i$ according to Eq.~\eqref{eqn:Z} is expensive.
We follow~\cite{mnih2013learning}, treating it as a constant
and estimating its value via Monte Carlo approximation:
\begin{equation}
Z \simeq  Z_i \simeq n E_j\left[\exp(\vv_j^T \vf_i / \tau)\right]
= \frac{n}{m} \sum_{k=1}^m \exp(\vv_{j_k}^T \vf_i /\tau),
\end{equation}
where $\{j_k\}$ is a random subset of indices.
Empirically, we find the approximation derived from initial batches sufficient to work well in practice.

NCE reduces the computational complexity from $O(n)$ to $O(1)$ per sample.
With such drastic reduction, our experiments
still yield competitive performance.

\subsection{Proximal Regularization}

\begin{figure}[hp]
	\centering
	\includegraphics[height=0.2\textheight]{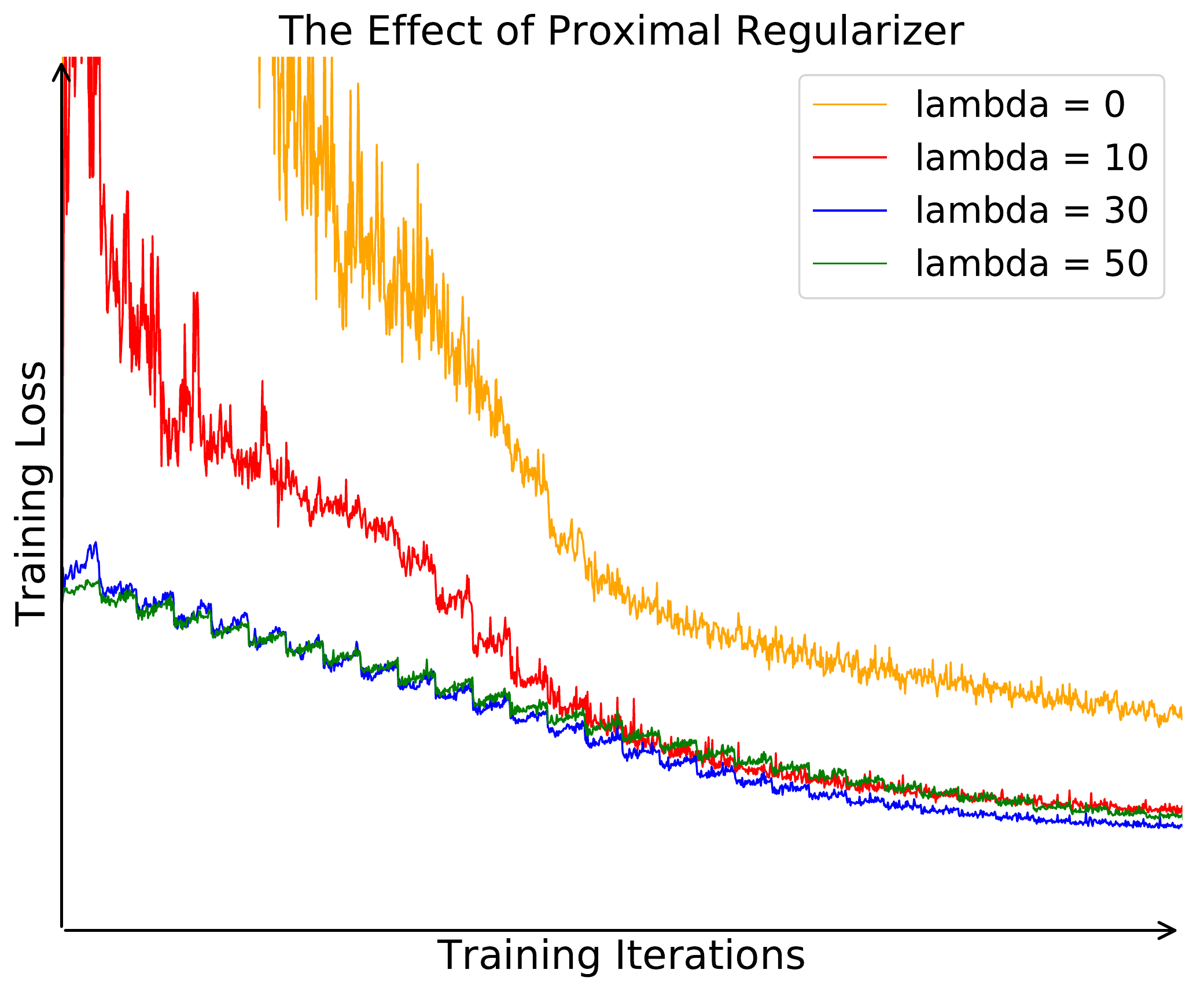}
	\caption{\small
		The effect of our proximal regularization.
		The original objective value oscillates a lot and converges very slowly,
		whereas the regularized objective has smoother learning dynamics.}
	\label{fig:proximal}
\end{figure}

Unlike typical classification settings where each class has many instances, we only have one instance per class.  During each training epoch,
each class is only visited once. Therefore, the learning process oscillates a lot from random sampling fluctuation.  We employ the proximal
optimization method~\cite{parikh2014proximal} and introduce an additional term
to encourage the smoothness of the training dynamics.
At current iteration $t$,
the feature representation for data $x_i$ is computed from the network $\vv_i^{(t)} = f_{\vtheta}(x_i)$.
The memory bank of all the representation are stored at previous iteration $V = \{\vv^{(t-1)}\}$.
The loss function for a positive sample from $P_d$ is:
\begin{equation}
- \log h(i, \vv_i^{(t-1)}) + \lambda \|\vv_i^{(t)} - \vv_i^{(t-1)}\|_2^2.
\end{equation}
%
As learning converges, the difference between iterations,
\ie~$\vv_i^{(t)} - \vv_i^{(t-1)}$, gradually vanishes, and
the augmented loss is reduced to the original one.
With proximal regularization, our final objective becomes:
\begin{align}
J_{NCE}(\vtheta)
= - & E_{P_d} \left[\log h(i, \vv_i^{(t-1)}) - \lambda\|\vv_i^{(t)} - \vv_i^{(t-1)}\|_2^2\right] \notag \\
   - m \cdot & E_{P_n} \left[\log (1 - h(i, \vv'^{(t-1)})) \right].
\end{align}
Fig.~\ref{fig:proximal} shows that, empirically, 
proximal regularization helps stabilize training, speed up convergence,
and improve the learned representation, with negligible extra cost.

\subsection{Weighted k-Nearest Neighbor Classifier}

To classify test image $\hat{x}$, we first compute its
feature $\hat{\vf} = f_{\vtheta}(\hat{x})$, and then
compare it against the embeddings of all the images in the memory bank, using
the cosine similarity $s_i = \cos(\vv_i, \hat{\vf})$. The top $k$
nearest neighbors, denoted by $\cN_k$,
would then be used to make the prediction via 
weighted voting. Specifically, the class $c$ would get a total weight
$w_c = \sum_{i \in \cN_k} \alpha_i \cdot 1(c_i = c) $.  Here, $\alpha_i$ is the contributing weight of neighbor $x_i$, which
depends on the similarity as $\alpha_i = \exp (s_i / \tau)$.
We choose $\tau = 0.07$ as in training and
we set $k=200$.



\section{Experiments}
\label{exp}

We conduct 4 sets of experiments to evaluate our  approach.
The first set is on CIFAR-10 to compare 
our non-parametric softmax with parametric softmax.
The second set is on ImageNet to compare
our method with other unsupervised learning methods.
The last two sets of experiments investigate two different tasks, semi-supervised
learning and object detection, to show the generalization ability of our learned feature representation.

\begin{table}[t]
	\setlength{\tabcolsep}{1.8pt}
	\centering
	\begin{tabular}{c|c|c}
		\Xhline{3\arrayrulewidth}
		Training / Testing & Linear SVM & Nearest Neighbor \\
		\Xhline{3\arrayrulewidth}
		Param Softmax & 60.3 & 63.0 \\
		\Xhline{3\arrayrulewidth}
		Non-Param Softmax & 75.4 & \textbf{80.8} \\
		\Xhline{3\arrayrulewidth}
		NCE $m=1$ & 44.3 & 42.5\\ \hline
		NCE $m=10$ & 60.2 & 63.4\\ \hline
		NCE $m= 512$ & 64.3   & 78.4\\ \hline
		NCE $m= 4096$ & 70.2  & \textbf{80.4}\\
		\Xhline{3\arrayrulewidth}
	\end{tabular}
	\caption{\small
	Top-1 accuracy on CIFAR10, by applying 
	linear SVM or kNN classifiers
		on the learned features.
	Our non-parametric softmax outperforms parametric softmax, and NCE provides close approximation as $m$ increases.
	}
	\label{exp:ablative}
\vspace{-3pt}
\end{table}

\subsection{Parametric vs. Non-parametric Softmax}
\label{p:np}

A key novelty of our approach is the non-parametric softmax function.
Compared to the conventional parametric softmax,
our softmax allows a non-parametric metric to transfer to supervised tasks.

We compare both parametric and non-parametric formulations on
CIFAR-10~\cite{krizhevsky2009learning}, a dataset with
$50,000$ training instances in $10$ classes. This size allows us to compute the non-parametric softmax in Eq.\eqref{eqn:np_softmax} without any
approximation. We use ResNet18 as the backbone network and its output features mapped into $128$-dimensional vectors.

We evaluate the classification effectiveness based on the learned feature representation.
A common practice~\cite{zhang2017split, doersch2015unsupervised, pathak2016context}
is to train an SVM on the learned feature over the training set, and 
to then classify test instances based on the feature extracted from the trained network.
In addition, we also use nearest neighbor classifiers to
assess the learned feature. The latter directly relies on the feature metric and
may better reflect the quality of the representation.

Table~\ref{exp:ablative} shows top-1 classification
accuracy on CIFAR10. On the features learned with parametric softmax, we obtain
accuracy of $60.3\%$ and $63.0\%$ with linear SVM and kNN classifiers respectively.  On the features learned with
non-parametric softmax, the accuracy 
rises to  $75.4\%$ and $80.8\%$ for the linear and nearest neighbour classifiers, a remarkable
$18\%$ boost for the latter.

We also study the quality of NCE approximating non-parametric softmax (Sec.~\ref{sub:nce}).
The approximation is controlled by $m$,
the number of negatives drawn for each instance.
With $m = 1$, the accuracy with kNN
drops significantly to $42.5\%$. As $m$ increases, the performance improves
steadily. When $m = 4,096$, the accuracy approaches that at $m = 49,999$ -- full form evaluation without any approximation.  This result provides assurance that NCE is an efficient approximation.

\subsection{Image Classification}

We learn a feature representation on ImageNet ILSVRC~\cite{russakovsky2015imagenet},
and compare our method with representative unsupervised learning methods.

\vspace{-9pt}
\paragraph{Experimental Settings.}
We choose design parameters via empirical validation.
In particular, we set temperature $\tau = 0.07$ and use NCE with $m = 4,096$ to balance performance and computing cost.
The model is trained for $200$ epochs using
SGD with momentum. The batch size is 256.
The learning rate is initialized to $0.03$,
scaled down with coefficient $0.1$ every 40 epochs after the first $120$ epochs.
Our code is available at: \url{http://github.com/zhirongw/lemniscate.pytorch}.

\vspace{-9pt}
\paragraph{Comparisons.}
We compare our method
with a randomly initialized network (as a lower bound) and
various unsupervised learning methods, including
self-supervised learning~\cite{doersch2015unsupervised,zhang2016colorful,noroozi2016unsupervised,zhang2017split},
adversarial learning~\cite{donahue2016adversarial}, and
Exemplar CNN~\cite{doersch2017multi}.
The split-brain autoencoder~\cite{zhang2017split} serves a strong baseline that represents the state of the art.
The results of these methods are reported with
AlexNet architecture ~\cite{krizhevsky2012imagenet} in their original papers,
except for exemplar CNN~\cite{dosovitskiy2014discriminative},
whose results are reported with ResNet-101~\cite{doersch2017multi}.
As the network architecture has a big impact on the performance,
we consider a few typical architectures:
AlexNet~\cite{krizhevsky2012imagenet},
VGG16~\cite{simonyan2014very},
ResNet-18,
and ResNet-50~\cite{he2015deep}.

\begin{table}[tp]
\setlength{\tabcolsep}{1.1pt}
\centering
\begin{tabular}{c|ccccc|c|c}
\Xhline{2\arrayrulewidth}
  \multicolumn{8}{c}{Image Classification Accuracy on ImageNet} \\
\Xhline{2\arrayrulewidth}
method  & conv1 & conv2 & conv3 & conv4 & conv5 & kNN & \#dim \\
\hline
Random & 11.6 & 17.1 & 16.9 & 16.3 & 14.1 & 3.5 & 10K \\
Data-Init~\cite{krahenbuhl2015data} & 17.5 & 23.0 & 24.5 & 23.2 & 20.6 & - & 10K\\
Context~\cite{doersch2015unsupervised} & 16.2 & 23.3 & 30.2 & 31.7 & 29.6 & - & 10K\\
Adversarial~\cite{donahue2016adversarial} & 17.7 & 24.5 & 31.0 & 29.9 & 28.0 & - & 10K\\
Color~\cite{zhang2016colorful} & 13.1 & 24.8 & 31.0 & 32.6 & 31.8 & - & 10K\\
Jigsaw~\cite{noroozi2016unsupervised} & 19.2 & 30.1 & 34.7 & 33.9 & 28.3 & - & 10K\\
Count~\cite{noroozi2017representation} & 18.0 & 30.6 & 34.3 & 32.5 & 25.7 & - & 10K \\
SplitBrain~\cite{zhang2017split} & 17.7 & 29.3 & 35.4 & 35.2 & 32.8 & 11.8 & 10K\\
\Xhline{2\arrayrulewidth}
Exemplar\cite{doersch2017multi} & & & 31.5 & & & - & 4.5K \\
\Xhline{2\arrayrulewidth}
Ours Alexnet & 16.8 & 26.5 & 31.8 & 34.1 & \textbf{35.6} & 31.3 & 128 \\
Ours VGG16 & 16.5 & 21.4 & 27.6 & 35.1 & \textbf{39.2} & 33.9 & 128 \\
Ours Resnet18 & 16.0 & 19.9 & 29.8 & 39.0 & \textbf{44.5} & \textbf{41.0} & 128\\
Ours Resnet50 & 15.3 & 18.8 & 24.9 & 40.6 & \textbf{54.0} & \textbf{46.5} & 128\\
\Xhline{2\arrayrulewidth}
\end{tabular}
\caption{\small
    Top-1 classification accuracy on ImageNet.
}
\label{exp:cls_imagenet}
\end{table}

\begin{table}[t]
	\setlength{\tabcolsep}{1.1pt}
	\centering
	\begin{tabular}{c|ccccc|c|c}
		\Xhline{2\arrayrulewidth}
		\multicolumn{8}{c}{Image Classification Accuracy on Places} \\
		\Xhline{2\arrayrulewidth}
		method & conv1 & conv2 & conv3 & conv4 & conv5 & kNN & \#dim  \\
		\hline
		Random &  15.7 & 20.3 & 19.8 & 19.1 & 17.5 & 3.9 & 10K \\
		Data-Init~\cite{krahenbuhl2015data} &  21.4 & 26.2 & 27.1 & 26.1 & 24.0 & - & 10K \\
		Context~\cite{doersch2015unsupervised} & 19.7 & 26.7 & 31.9 & 32.7 & 30.9 & - & 10K \\
		Adversarial~\cite{donahue2016adversarial} &  17.7 & 24.5 & 31.0 & 29.9 & 28.0 & - & 10K \\
		Video~\cite{wang2015unsupervised} & 20.1 & 28.5 & 29.9 & 29.7 & 27.9 & - & 10K \\
		Color~\cite{zhang2016colorful} & 22.0 & 28.7 & 31.8 & 31.3 & 29.7 & - & 10K \\
		Jigsaw~\cite{noroozi2016unsupervised} & 23.0 & 32.1 & 35.5 & 34.8 & 31.3 & - & 10K\\
		SplitBrain~\cite{zhang2017split} & 21.3 & 30.7 & 34.0 & 34.1 & 32.5 & 10.8 & 10K \\
		\Xhline{2\arrayrulewidth}
		Ours Alexnet & 18.8 & 24.3 & 31.9 & \textbf{34.5} & 33.6 & 30.1 & 128 \\
		Ours VGG16 & 17.6 & 23.1 &  29.5 & 33.8 & \textbf{36.3} & 32.8 & 128 \\
		Ours Resnet18 & 17.8 & 23.0 & 30.1 & 37.0 & \textbf{38.1} & \textbf{38.6} & 128\\
		Ours Resnet50 & 18.1 & 22.3 & 29.7 & 42.1 & \textbf{45.5} & \textbf{41.6} & 128\\
		\Xhline{2\arrayrulewidth}
	\end{tabular}
	\caption{\small
        Top-1 classification accuracy on Places,  based directly on features learned on ImageNet, without any fine-tuning.
    }
	\label{exp:cls_places}
\end{table}

We evaluate the performance with two different protocols:
(1) Perform linear SVM on the intermediate features from
\texttt{conv1} to \texttt{conv5}. Note that 
there are also corresponding layers in VGG16 and ResNet~\cite{simonyan2014very,he2015deep}.
(2) Perform kNN on the output features.
%
Table~\ref{exp:cls_imagenet} shows that:

\vspace{-7pt}
\begin{enumerate}[leftmargin=*,labelindent=0pt,itemsep=0pt]
\item With AlexNet and linear classification on intermediate features,
our method achieves an accuracy of $35.6\%$,
outperforming all baselines, including the state-of-the-art.
Our method can readily scale up to deeper networks. As we move from
AlexNet to ResNet-50, our accuracy is raised to $54.0\%$,
whereas the accuracy with
exemplar CNN~\cite{doersch2017multi} 
is only $31.5\%$ even with ResNet-101.

\begin{figure}[t]
	\centering
	\includegraphics[height=140pt]{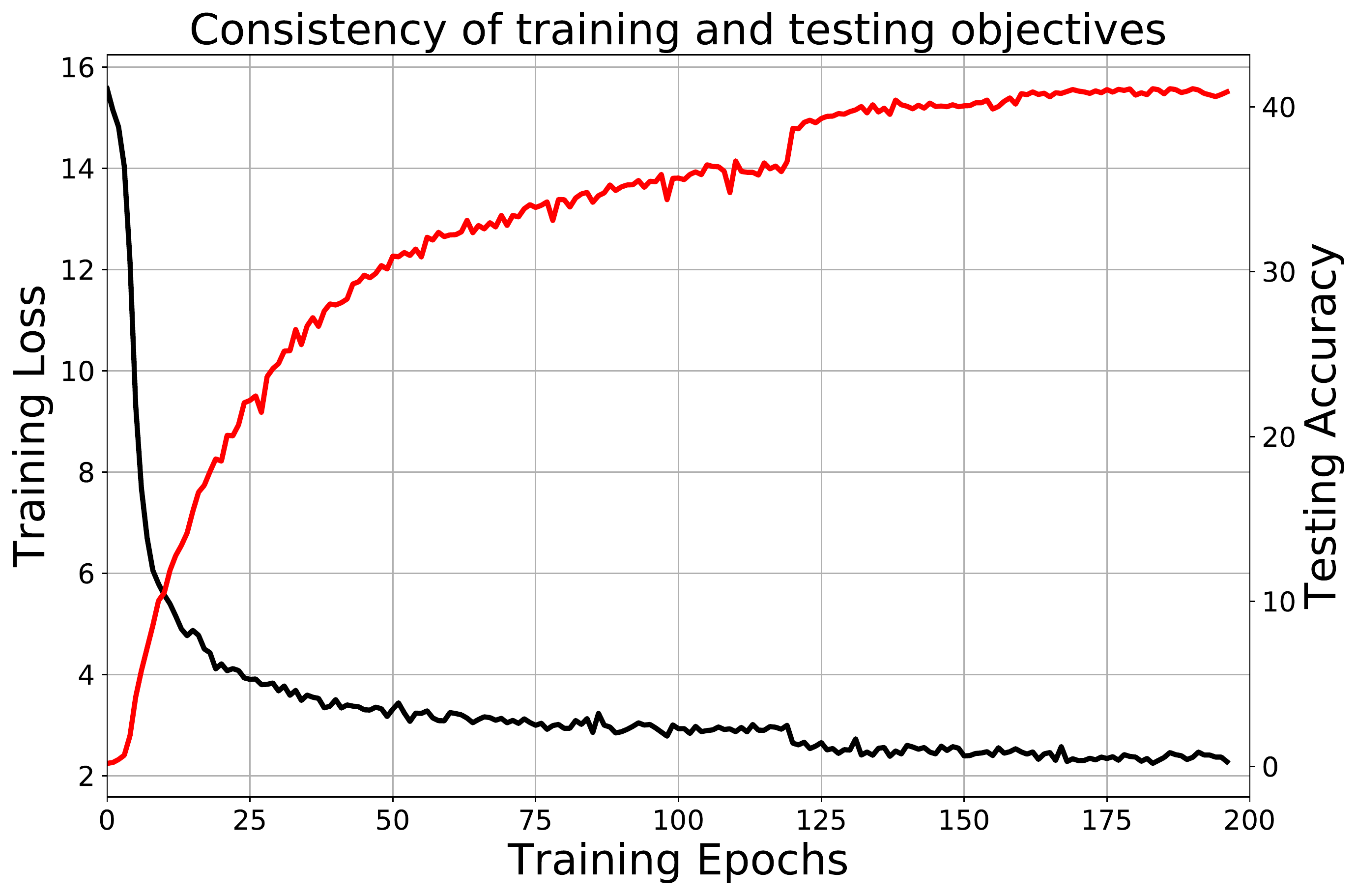}
	\caption{\small
		Our kNN testing accuracy on ImageNet continues to improve as the training loss decreases, demonstrating 
		that our unsupervised learning objective captures apparent similarity which aligns well with the semantic annotation of the data.
	}
	\label{fig:consistency}
\vspace{-6pt}
\end{figure}

\item Using nearest neighbor classification on the final 128 dimensional features, our method achieves $31.3\%$, $33.9\%$, $41.0\%$ and $46.5\%$ accuracies
with AlexNet, VGG16, ResNet-18 and ResNet-50,
not much lower than the linear classification results, demonstrating that our learned feature induces a reasonably good metric.
As a comparison, for Split-brain, the accuracy drops to $8.9\%$ with nearest neighbor classification on \texttt{conv3} features,
and to $11.8\%$ after projecting the features to 128 dimensions.

\item With our method, the performance gradually increases
as we examine the learned feature representation from earlier to later layers, which is generally desirable.
With all other methods,
the performance decreases beyond \texttt{conv3} or \texttt{conv4}.

\item It is important to note that the features from
intermediate convolutional layers can be over $10,000$ dimensions.
Hence, for other methods, using the features from the \emph{best-performing}
layers can incur significant storage and computation costs.
Our method produces a $128$-dimensional representation at the last layer,
which is very efficient to work with. The encoded features of all $1.28M$
images in ImageNet only take about $600$ MB of storage.
Exhaustive nearest neighbor search over this dataset only takes $20$ ms per image on a Titan X GPU.
\end{enumerate}

\begin{figure*}[t]
	\centering
	\includegraphics[width=0.98\linewidth]{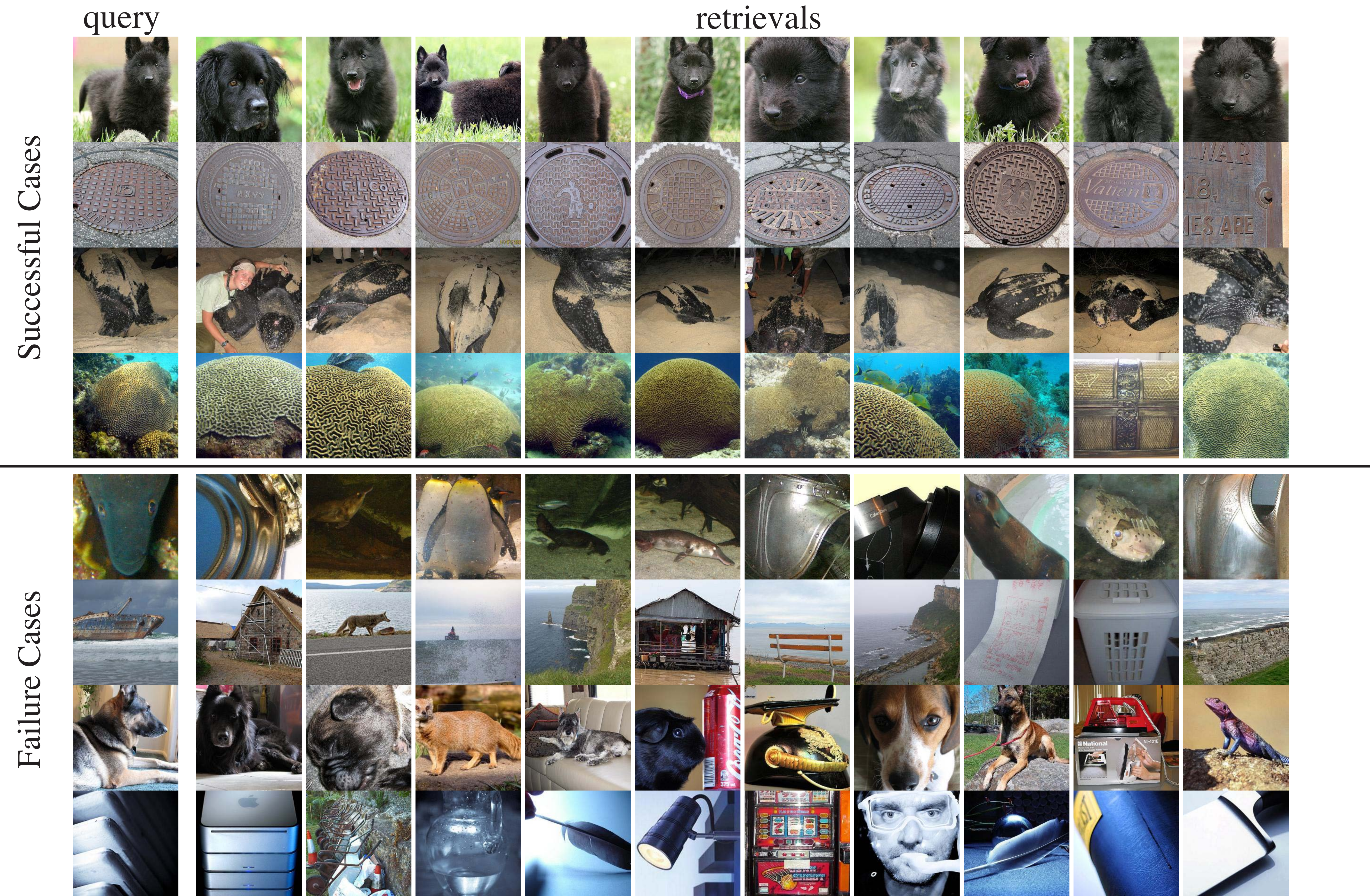}
	\caption{\small
		Retrieval results for example queries.
		The left column are queries from the validation set, while
		the right columns show the $10$ closest instances from the training set.
		The upper half shows the best cases. The lower half shows the worst cases.
	}
	\label{fig:retrieval}
\vspace{-3pt}
\end{figure*}

\vspace{-18pt}
\paragraph{Feature generalization.}
We also study how the learned feature representations can
generalize to other datasets. With the same settings, we conduct another
large-scale experiment on \emph{Places}~\cite{zhou2014learning}, a large
dataset for scene classification, which contains $2.45M$ training images
in $205$ categories.
In this experiment, we directly use the feature extraction networks trained
on ImageNet \emph{without finetuning}. Table~\ref{exp:cls_places} compares
the results obtained with different methods and under different evaluation policies. Again, with linear classifier on \texttt{conv5} features,
our method achieves competitive performance of
top-1 accuracy $34.5\%$ with AlexNet, and 
 $45.5\%$ with ResNet-50.
With nearest neighbors on the last layer which is much smaller than intermediate layers, we achieve an accuracy of $41.6 \%$ with ResNet-50. These results show
remarkable generalization ability of the representations learned using our method.

\vspace{-9pt}
\paragraph{Consistency of training and testing objectives.}
Unsupervised feature learning is difficult because the training objective is agnostic about the testing objective. 
A good training objective should be reflected in consistent improvement in the 
testing performance.  We investigate the relation between the training loss
and the testing accuracy across iterations.
Fig.~\ref{fig:consistency} shows that our testing accuracy continues to improve 
as training proceeds, with no sign of overfitting.
It also suggests that better optimization of the training objective
may further improve our testing accuracy.

\vspace{-9pt}
\paragraph{The embedding feature size.}
We study how the performance changes as  we vary the embedding size from 32 to 256.
Table~\ref{table:feat_size} shows that the performance increases from 32, plateaus at 128, and appears to saturate towards 256.

\begin{table}[t]
\centering
\begin{tabular}{c|c|c|c|c}
\Xhline{2\arrayrulewidth}
embedding size & $32$ & $64$ & $128$ & $256$ \\
\Xhline{2\arrayrulewidth}
top-1 accuracy & 34.0 & 38.8 & 41.0 & 40.1\\
\Xhline{2\arrayrulewidth}
\end{tabular}
\caption{\small
Classification performance on ImageNet with ResNet18 for different embedding feature sizes.
}
\label{table:feat_size}
\vspace{-9pt}
\end{table}

\vspace{-9pt}
\paragraph{Training set size.}
To study how our method scales with the data size,
we train different representations with various proportions of ImageNet data,
and evaluate the classification performance on
the full labeled set using nearest neighbors.
Table~\ref{table:data_scale} shows that our feature learning method benefits from larger
training sets, and the testing accuracy improves as the training set grows.
This property is crucial for successful unsupervised learning, as there is no shortage of unlabeled data in the wild.

\begin{table}[t]
	\centering
	\begin{tabular}{c|c |c|c|c|c}
		\Xhline{2\arrayrulewidth}
		training set size & $0.1\%$ & $1\%$ & $10\%$ & $30\%$ & $100\%$ \\
		\Xhline{2\arrayrulewidth}
		accuracy & 3.9 & 10.7 & 23.1 &  31.7 & 41.0 \\
		\Xhline{2\arrayrulewidth}
	\end{tabular}
	\caption{\small
		Classification performances
		trained on different amount of training set
with ResNet-18.
	}
	\label{table:data_scale}
\vspace{-9pt}
\end{table}

\vspace{-11pt}
\paragraph{Qualitative case study.}
To illustrate the learned features,
Figure~\ref{fig:retrieval} shows the results of image retrieval using the learned
features.
The upper four rows show the best cases where all top 10 results
are in the same categories as the queries.
The lower four rows show the worst cases where none of the top 10
are in the same categories. However, even for the failure cases, the retrieved images are still visually similar to the queries, a testament to the power of our unsupervised learning objective.

\subsection{Semi-supervised Learning}

We now study how the learned feature extraction network can benefit other tasks, and whether it can provide a good basis for transfer learning to other tasks.
A common scenario that can benefit from unsupervised learning is when we have a large amount of data of which only a small fraction are labeled.
A natural semi-supervised learning approach is to first learn from the big unlabeled data and then fine-tune the model
on the small labeled data.

We randomly choose a subset of ImageNet as
labeled and treat others as unlabeled.
We perform the above semi-supervised learning and measure the classification accuracy on the validation set.
In order to compare with \cite{larsson2017colorization}, we report the top-5 accuracy here.

We compare our method with three baselines:
(1) \emph{Scratch}, \ie~fully supervised training on the small labeled subsets,
(2) \emph{Split-brain}~\cite{zhang2017split} for pre-training,
and (3) \emph{Colorization}~\cite{larsson2017colorization} for pre-training.
Finetuning on the labeled subset takes $70$ epochs with initial learning
rate $0.01$ and a decay rate of 10 every 30 epochs.
We vary the proportion of labeled subset 
from $1\%$ to $20\%$ of the entire dataset.

\begin{figure}[t]
	\centering
	\includegraphics[height=180pt]{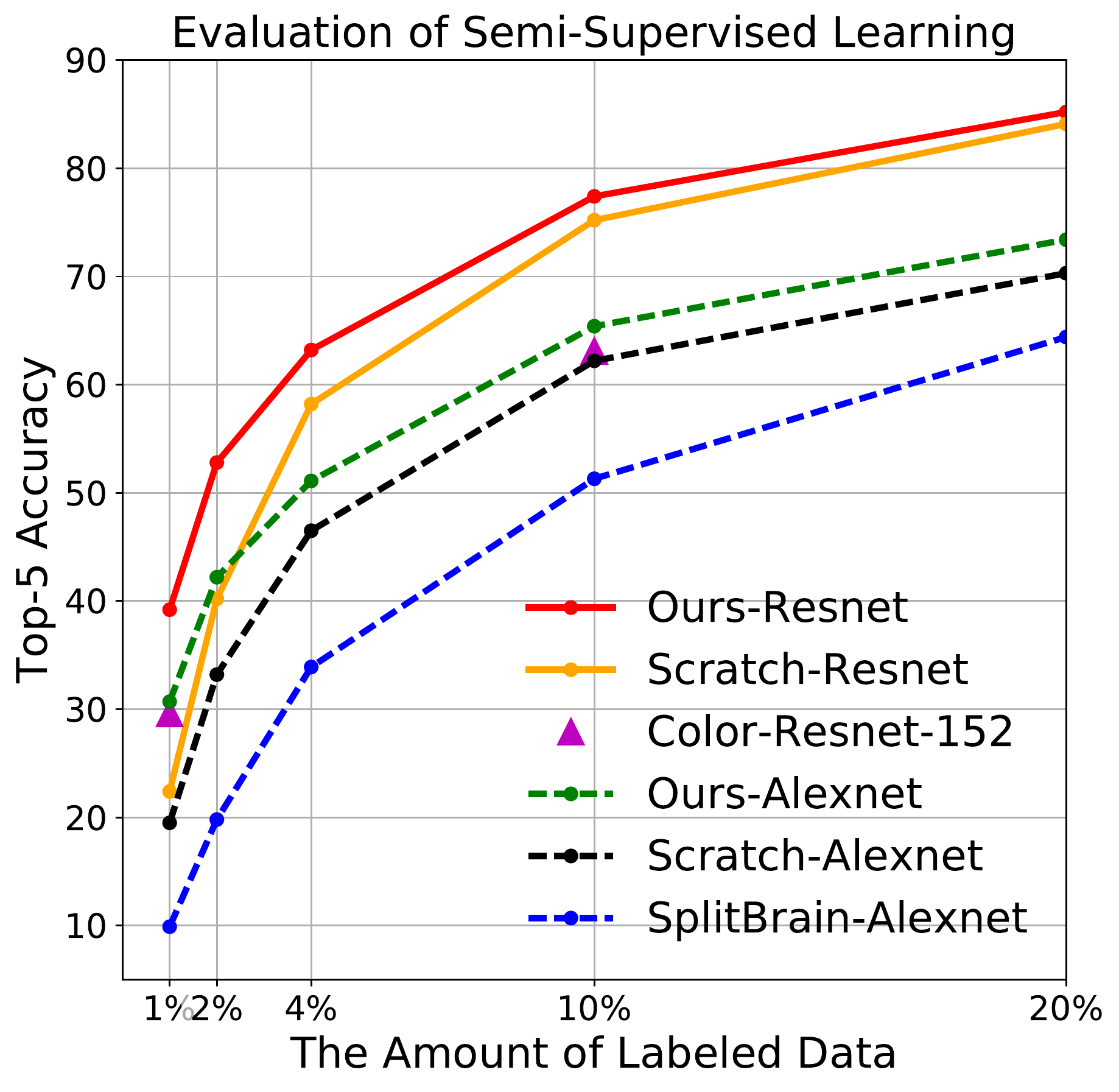}
	\caption{\small
		Semi-supervised learning results on ImageNet with an increasing fraction of labeled data ($x$ axis).  Ours are consistently and significantly better.
		Note that the results for colorization-based pretraining are from a deeper ResNet-152 network~\cite{larsson2017colorization}.
	}
	\label{fig:semi}
\vspace{-6pt}
\end{figure}

Fig.~\ref{fig:semi} shows
that our method significantly outperforms all other approaches, and ours is the only one outperforming supervised learning from limited labeled data.
When only $1\%$ of data is labeled, we outperform by a large $10\%$ margin,
demonstrating that our feature learned
from unlabeled data is effective for task adaptation.


\subsection{Object Detection}

To further assess the generalization capacity of the learned features, we
transfer the learned networks to the new task of object detection on PASCAL VOC 2007~\cite{everingham2010pascal}.
Training object detection model from scratch is often difficult, and a prevalent practice is to pretrain the underlying CNN on ImageNet and
fine-tune it for the detection task.

We experiment with
Fast R-CNN~\cite{girshick2015fast} with AlexNet and VGG16 architectures,
and Faster R-CNN~\cite{ren2015faster} with ResNet-50.
When fine-tuning Fast R-CNN, the learning rate is initialized to $0.001$ and
scaled down by $10$ times after every $50K$ iterations.
When fine-tuning AlexNet and VGG16, we follow the standard practice, fixing
the conv1 model weights.
When fine-tuning Faster R-CNN, we fix
the model weights below the 3rd type of residual blocks,
only updating the layers above and freezing all batch normalization layers.
We follow the standard pipeline for finetuning and do not use the rescaling method proposed in~\cite{doersch2015unsupervised}.
We use the standard trainval set in VOC 2007 for training and  testing.

We compare three settings:
1) directly training from scratch (lower bound),
2) pretraining on ImageNet in a supervised way (upper bound), and
3) pretraining on ImageNet or other data using various unsupervised methods.

Table~\ref{exp:det} lists detection performance in terms of mean average precision (mAP).
With AlexNet and VGG16, our method achieves an mAP of $48.1\%$ and $60.5\%$,
on par with the state-of-the-art unsupervised methods.
With Resnet-50, our method achieves an mAP of $65.4\% $,
surpassing all existing unsupervised learning approaches.
It also shows that our method scales well as the network gets deeper.
There remains a significant gap of $11\%$ to be narrowed towards mAP $76.2\%$ from supervised pretraining.

\begin{table}[t]
\setlength{\tabcolsep}{1.8pt}
~~~~~~~
\begin{minipage}{0.3\linewidth}
\begin{tabular}[c]{c|c}
\Xhline{2\arrayrulewidth}
\multicolumn{1}{c|}{Method}                                          & mAP    \\ \hline
\multicolumn{1}{c|}{AlexNet Labels$\dagger$}                         & $56.8$ \\
\multicolumn{1}{c|}{Gaussian}                                        & $43.4$ \\
\multicolumn{1}{c|}{Data-Init~\cite{krahenbuhl2015data}}     & $45.6$ \\
\multicolumn{1}{c|}{Context~\cite{doersch2015unsupervised}}   & $\mathbf{51.1}$ \\
\multicolumn{1}{c|}{Adversarial~\cite{donahue2016adversarial}}    & $46.9$ \\
\multicolumn{1}{c|}{Color~\cite{zhang2016colorful}}           & $46.9$ \\
\multicolumn{1}{c|}{Video~\cite{wang2015unsupervised}}         & $47.4$ \\
\multicolumn{1}{c|}{Ours Alexnet}                             &   $48.1$ \\
\Xhline{2\arrayrulewidth}
\end{tabular}
\end{minipage}
~~~~~~~~~
\begin{minipage}{0.3\linewidth}
\begin{tabular}[c]{c|c}
\Xhline{2\arrayrulewidth}
\multicolumn{1}{c|}{Method}                                          & mAP    \\ \hline
\multicolumn{1}{c|}{VGG Labels$\dagger$}                         & $67.3$ \\
\multicolumn{1}{c|}{Gaussian}                                        & $39.7$ \\
\multicolumn{1}{c|}{Video~\cite{wang2015unsupervised}}     & $60.2$ \\
\multicolumn{1}{c|}{Context~\cite{doersch2015unsupervised}}   & $61.5$ \\
\multicolumn{1}{c|}{Transitivity~\cite{wang2017transitive}}   & $\mathbf{63.2}$ \\
\multicolumn{1}{c|}{Ours VGG}                                      & $60.5$ \\
\hline
\hline
\multicolumn{1}{c|}{ResNet Labels$\dagger$} & $76.2$ \\
\multicolumn{1}{c|}{Ours ResNet}                                            & $\mathbf{65.4}$ \\
\Xhline{2\arrayrulewidth}
\end{tabular}
\end{minipage}
\caption{Object detection performance on PASCAL VOC 2007 test,
	 in terms of mean average precision (mAP), for 
	supervised pretraining methods (marked by $\dagger$),
 existing unsupervised methods, and our method.
}\label{exp:det}
\vspace{-9pt}
\end{table}


\section{Summary}

We present an unsupervised feature learning approach by maximizing distinction between instances via a novel non-parametric softmax formulation. It is motivated by the observation that supervised learning results in apparent image similarity.
Our experimental results show that our method outperforms the state-of-the-art on image classification on ImageNet and Places,
with a compact $128$-dimensional representation that scales well with more data and deeper networks.
It also delivers competitive generalization results on semi-supervised learning and object detection tasks. %
%

{\bf Acknowledgements.}
This work was supported in part by Berkeley Deep Drive,  Big Data Collaboration Research grant from SenseTime Group (CUHK Agreement No. TS1610626), and the General Research Fund (GRF) of Hong Kong (No. 14236516).

{\small
\bibliographystyle{ieee}
\bibliography{egbib}
}

\end{document}